\documentclass[sigconf,fontsize=9pt]{acmart}

\usepackage{booktabs} 
\usepackage[utf8]{inputenc} 
\usepackage[T1]{fontenc}    
\usepackage{lmodern}
\usepackage{hyperref}       
\usepackage{url}            
\usepackage{booktabs}       
\usepackage{amsfonts}       
\usepackage{nicefrac}       
\usepackage{microtype}      
\usepackage{graphicx}
\usepackage{wrapfig}
\usepackage{enumitem}
\usepackage{amsmath}
\usepackage{amssymb}
\usepackage{multirow}
\usepackage{makecell}
\usepackage{xcolor, colortbl}

\renewcommand\footnotetextcopyrightpermission[1]{} 
\definecolor{Gray}{gray}{0.85}
\newcolumntype{a}{>{\columncolor{Gray}}c}
\begin{document}
\title[Distill-Net]{Distill-Net: Application-Specific Distillation of \\Deep Convolutional Neural Networks for\\ Resource-Constrained IoT Platforms}

\author{Mohammad Motamedi}
\orcid{0000-0003-0120-8738}
\affiliation{%
  \institution{University of California, Davis}
  \streetaddress{One Shields Avenue}
  \city{Davis}
  \state{CA}
  \postcode{95616}
  \country{USA}}
\email{mmotamedi@ucdavis.edu}
\author{Felix A. Portillo}
\affiliation{%
  \institution{University of California, Davis}
\streetaddress{One Shields Avenue}
\city{Davis}
\state{CA}
\postcode{95616}
\country{USA}}
\email{faportillo@ucdavis.edu}
\author{Daniel Fong}
\affiliation{%
  \institution{University of California, Davis}
\streetaddress{One Shields Avenue}
\city{Davis}
\state{CA}
\postcode{95616}
\country{USA}}
\email{dfong@ucdavis.edu}
\author{Soheil Ghiasi}
\affiliation{%
  \institution{University of California, Davis}
\streetaddress{One Shields Avenue}
\city{Davis}
\state{CA}
\postcode{95616}
\country{USA}}
\email{ghiasi@ucdavis.edu}

\begin{abstract}
Many Internet-of-Things (IoT) applications demand fast and accurate understanding of a few key events in their surrounding environment. Deep Convolutional Neural Networks (CNNs) have emerged as an effective approach to understand speech, images, and similar high dimensional data types. Algorithmic performance of modern CNNs, however, fundamentally relies on learning class-agnostic hierarchical features that only exist in comprehensive training datasets with many classes. As a result, fast inference using  CNNs trained on such datasets is prohibitive for most resource-constrained IoT platforms. 
To bridge this gap, we present a principled and practical methodology for distilling a complex modern CNN that is trained to effectively recognize many different classes of input data into an application-dependent essential core that not only recognizes the few classes of interest to the application accurately, but also runs efficiently on platforms with limited resources. Experimental results confirm that our approach strikes a favorable balance between classification accuracy (application constraint), inference efficiency (platform constraint), and productive development of new applications (business constraint).
\end{abstract}

%
%

%
%

\maketitle

\renewcommand{\shortauthors}{M. Motamedi et al.}

\section{Introduction}
Deep Convolutional Neural Networks (CNN) have demonstrated remarkable performance in pattern recognition and classification tasks on high dimensional data, such as visual recognition and speech understanding. State of the art CNNs need to be trained on comprehensive labeled datasets, which include a large number of classes and many examples per class. Such CNNs typically involve billions of operations on millions of parameters, hindering their utilization in IoT applications that demand fast inference on resource-constrained execution platforms. 

\begin{figure}
	\includegraphics[width=7cm]{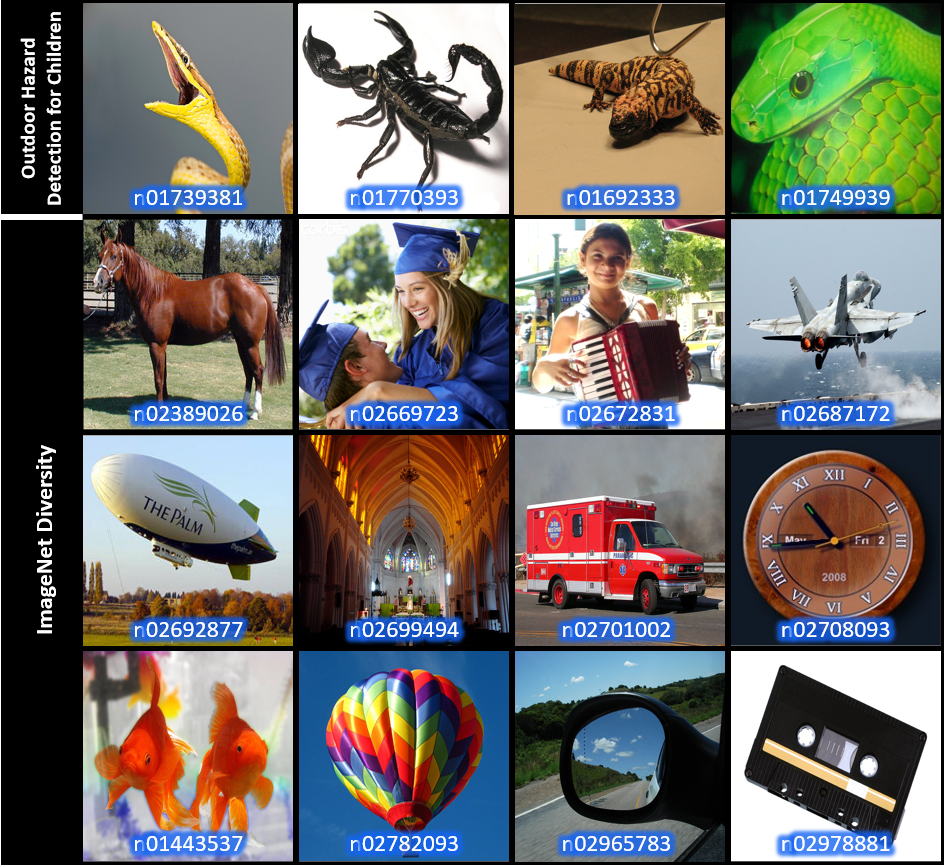}
	\caption{\small Samples from ILSVRC dataset~\cite{ILSVRC15}. The first row shows images from classes that are essential for an outdoor hazard detection application, whereas other rows present samples from just a few classes, out of hundreds of classes, that are not useful for this task. The code of the class to which each image belongs is shown.}
	\label{Fig:ilsvrcdiversity}
\end{figure}

We observe that a significant subset of IoT applications~\cite{bandyopadhyay2011internet} require recognition of a fairly small number of classes, compared to the number of classes that exist in commonly used labeled training datasets. For example, disaster response drones or security cameras tend to require detection of a few key classes, such as person, tree, animal, road, obstacle and such, while the ILSVRC dataset \cite{ILSVRC15} that is used to train many recent CNNs includes more than 1000 classes, such as various breeds of dogs, types of bugs, and many exotic animals and uncommon objects. Figure~\ref{Fig:ilsvrcdiversity} shows the diversity of image classes in ILSVRC dataset~\cite{ILSVRC15}. The first row includes samples from classes that are beneficial for an outdoor hazard detection application for children. Other rows show images from just a few, out of hundreds, of classes that are not useful for this application. IoT systems are mission-oriented in nature, and thus, require recognition of classes that matter to their application. For many typical applications, such as those arising in precision agriculture, smart environment, navigation, and security~\cite{atzori2010internet}, the number of classes of interest is limited.

Given that large labeled datasets are expensive to generate, it is unrealistic to expect that for every new application, new labeled training data is available, and a new model can be constructed and trained from scratch. A key open question, thus, has to do with a practical approach to build application-specific CNNs that offer accurate, from a classification viewpoint, and efficient, from an execution perspective, mission-specific inference on resource-constrained IoT platforms. In this paper, we propose a technique towards addressing this very question. 

On the surface, the problem may seem misleadingly trivial. It may appear that only the training samples relevant to the target application could be utilized to train a simpler custom CNN.  This approach fails to produce CNNs that meet classification accuracy constraints: CNNs are effective in complex tasks, in part because they manage to learn class-agnostic features that exists only in comprehensive datasets. Training a network using a small subset of application-specific examples deprives the network of such features, which in turn, harms its accuracy. In other words, negative examples contain significant information, and removing them from the training dataset is detrimental to classification accuracy. 

Another naive approach to the problem would be to start with a state of the art trained CNN, and replace its final-stage classifier with an application-specific classifier that detects only a few classes of interest to the application. This approach certainly meets the accuracy requirement. However, it results in a CNN that is computationally very similar to the original model, and thus, does not yield fast inference on IoT platforms. Note that the overwhelming majority of computational workload of a CNN does not occur in its classifier stage \cite{szegedy2015going}. Such a process demands a follow up fine-tuning which includes multiple epochs. The required time for \textbf{each} epoch is five times larger than the required time for the entire distillation~\cite{jia2014caffe}.

Starting with a state of the art trained CNN and the comprehensive publicly-available dataset that was used for its training, we aim to systematically develop a custom CNN that meets the two inference efficiency and classification accuracy constraints, albeit on a small subset of classes. In our proposed methodology, IoT designers specify the classes of interest to the application. Subsequently, we analyze the input CNN, and distill it into its essential core with respect to the target application. Experimental results show that Distill-Nets at most drop the accuracy by 1\% for classes of interest to the application, while running significantly faster.

\section{Landscape and Contributions}
In theory, one could design, train and optimize a CNN to meet the requirements of a given IoT application.  In practice, this approach faces several challenges and shortcomings:\\
(1) Time-consuming nature of training deep networks:
Designing and training a new industrial-strength CNN for every new IoT application increases time to market. Such a process requires many rounds of trial-and-error, training, and optimization. Each round of training can take from a day to a week depending on the available hardware and the CNN architecture. In contrast, the proposed distillation approach can finish the task in a matter of minutes.\\
(2) Inherent difficulty of developing competitive models: While embedded system developers are experts in various aspects of designing time-predictable, and energy-efficient applications for resource-constrained IoT platforms, they are not necessarily proficient in developing and training competitive deep neural networks.\\
(3) Considerable cost of computational infrastructure to support the training: The process of designing and training a new CNN is computationally intensive and demands significant investment in computing infrastructure.\\
(4) Data privacy: The proposed distillation algorithm can be performed on a mediocre IoT platform (e.g., Google Nexus 6P) making it feasible to tailor trained CNNs to users' private data. However, retraining a CNN in such a case requires the data to be uploaded to a cloud which may compromise users' privacy.\\
(5) Dynamic mission assignment: In the training-based approaches, re-purposing a CNN to identify a new set of classes requires it to be retrained. However, since the distillation process we are proposing preserves the knowledge (i.e., weights) of all classes, redeploying a CNN for a new set of image objectives becomes quick and feasible. Distillation merely creates guidance bitmaps for the thread launcher to specify neurons whose computation can be skipped. That is, it preserves the CNN structure, and the corresponding weights (i.e., the knowledge of the original CNN is maintained). As a result, reassigning the purpose of an embedded application with a distilled CNN is simply the matter of discarding or updating the bitmaps.

Our study is performed under two constraints: First, in all of the experiments, the distillation algorithm is targeted to lose at most 1\% of the classification accuracy. In practice, in more than 60\% of our experiments we achieved less than 0.1\% loss in the classification accuracy.

Second, IoT devices are required to upload their private or proprietary data to a computation cloud, if they need a CNN to be tailored to their specific application. We believe that users' privacy protection must be considered an integral part of any IoT solution. The proposed distillation is designed under this constraint. In section~\ref{sec:Experimental Results}, we show that  a mediocre IoT device can run the algorithm on users' private data in a reasonable time budget.

The main contributions of this paper are: First, for the first time, we propose an approach to tailor a CNN for an IoT application whose objects of interest are a fraction of categories that the original CNN was designed to classify. Second, the proposed approach requires no additional \textbf{data}, training or fine-tuning, thus, addressing practical concerns, such as resource-aware inference, dynamic updates, and productive development.

Note that the term ``distillation'' is coincidentally used in some other articles such as \cite{hinton2015distilling,chen2017learning} to refer to a CNN training approach that utilizes the knowledge of an ensemble of trained CNNs in addition to the training set. It is essential to emphasize that while this paper uses the same term, it targets a substantially different problem.
In the distillation algorithm proposed in this article, we offer a solution for tailoring a CNN to an IoT application whose objects of interest are a subset of categories that the original CNN was designed to classify.
\section{Convolutional Neural Networks}
State-of-the-art CNNs are capable of making mostly correct classifications about images with invariance to the shape, size, rotation, or position of the desired object in a scene \cite{krizhevsky2012imagenet,szegedy2015going}. The key feature of CNNs is the convolutional layers, which consist of sets of trainable 3D filter-banks. These filter-banks extract features from Input Feature Maps (IFMs) to produce sets of Output Feature Maps (OFMs). 
By stacking multiple layers, a CNN can extract more abstract information from simple edges and colors to entire objects and patterns as the network gets deeper. However, having too many layers increases the capacity of the neural network (i.e., larger number of parameters), which makes it prone to over-fitting.

In essence, each 3D filter-bank is a neuron with a restricted receptive field. In the rest of this paper, the terms filter-bank and neuron are used interchangeably to convey the message clearly.
\section{Problem Statement}
Let us assume that a given CNN, $\Psi$, is trained using a large dataset whose data belongs to $\alpha$ distinct classes which are shown by set $A$ in Equation~(\ref{eq:setAB}). Let us also assume that for a particular IoT application, only $\beta$ classes that are shown in set $B$ are required, such that, $B \subset A$. The question is, can we use $\Psi$ to create a smaller CNN  that is adequate for classifying members of set $B$? 
	\begin{equation}
		\label{eq:setAB}
		\small A = \{a_1, a_2, \cdots, a_{\alpha}\},~B = \{b_1, b_2, \cdots, b_\beta\}
\end{equation}
The original CNN can serve the purpose. However, for each inference, it extracts all the features that are required for understanding members of set $A$, some of which will never be used to classify images of set $B$. In order to improve the execution time and energy consumption, we are interested in finding neurons that solely react to members of set $\overline{B}$ and removing them. $\overline{B}$ is the complement of set $B$ where set $A$ is the universe.
\section{CNN-based Feature Extraction}
One of the most important characteristics of CNNs that distinguishes them from other image recognition approaches is their ability in automatic feature extraction.  Given a large labeled dataset and sufficient computing resources, a CNN with an appropriate capacity can automatically extract adequate feature-sets that can be used for image classification. Features are learned in a layer-based fashion from the shallowest layer to the deepest one. In the very first layer, basic feature extractors such as line, curve, and dot detectors form. Even though such basic features do not directly contribute to the classification step, they have a very critical role in improving the final accuracy as they are the building blocks based on which feature extractors in subsequent layers form.

In shallow layers, features are simple and structurally related to the input. In contrast, features that are obtained in deep layers are more complicated and semantically related to the input. The functionality of neurons that generate such features (semantically related) is location-agnostic. That is, the stimulus that makes them fire is the existence of a pattern not its location.

\begin{figure}
	\includegraphics[width=\linewidth]{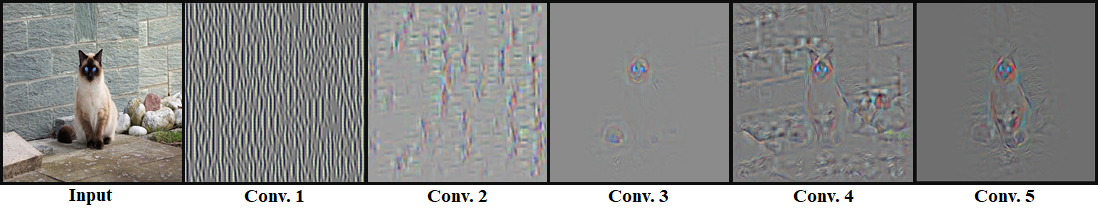}
	\caption{\small Strongest feature map for each layer of AlexNet~\cite{krizhevsky2012imagenet}. As we move towards deeper layers, features are more complex, exaggerate discriminative parts, and have a greater invariance.}
	\label{Fig:AlexNetVisualization}
\end{figure}

We used Deconvolutional Neural Networks \cite{zeiler2014visualizing,tf_cnnvis} to visualize feature maps for different layers of AlexNet~\cite{krizhevsky2012imagenet}. Figure \ref{Fig:AlexNetVisualization} illustrates the strongest feature map per each layer of the CNN. For this specific example, rich features start forming in convolutional layer \#3. In subsequent layers, CNN improves the quality of the features to make them even more distinctive for different classes in the classification task. Features which are extracted in deep layers intensify discriminative parts of objects. In this example, AlexNet mainly uses the face, eyes, and nose, for distinguishing a cat. 

In this paper, we use Class-Agnostic Features (CAFs) and Class-Dependent Features (CDFs) terms to refer to features that are extracted in shallow layers and deep layers, respectively. The former exist in all images and the latter exist in images that belong to one or a few specific classes. Likewise, Class-Agnostic Neurons (CANs) and Class-Dependent Neurons (CDNs) are terms used for neurons that generate CAFs and CDFs, respectively. CNNs utilize all of the labeled data (from all classes) to train CANs. However, they are limited to the data of each class for training its CDNs. 

When the labeled data that is available for training a particular CNN is inadequate, it is possible to use another dataset, which includes the same type of data, to train CANs. Subsequently, the original dataset must be used to fine-tune CANs and train CDNs. 

\section{Neuroactivity Analysis in CNNs}
Since Output Feature Maps (OFMs) store neurons' outputs, their sparsity is a good measure of neuron activities in each layer. Neuron activity is the reaction of a neuron to a particular input, i.e., whether it fires and if so, how strongly it does. In the first part of this section, we propose an approach for quantifying the sparsity of an OFM. We hypothesize that OFMs which hold outputs of CANs are less sparse compared to those that store outputs of CDNs since CANs fire more frequently. In the second part of this section, we check the validity of this hypothesis in GoogLeNet~\cite{szegedy2015going}.

\subsection{Dormant Neurons Density}
Computation that a neuron performs on a tile $t$ of IFMs is shown in Equation (\ref{eq:neuron_computation}), in which $f^t$, $K$, $N$, and $b$ are neuron's output for the tile $t$, kernel size, number of IFMs, and the bias value, respectively. In this equation, $X^t$ is a matrix including a tile of the IFM, i.e., the neuron activities from the previous layer, $W$ is a matrix that contains parameters of a neuron, and the ``+'' superscript shows the positive part of the function.

If Equation~(\ref{eq:simp_cond}) holds (linear flattening), Equation~(\ref{eq:neuron_computation}) can be rewritten in a simplified form which is shown in Equation~(\ref{eq:simplified_conv}). 

We define $X''$ to be a sorted equivalent of $X'$ whose values are in a descending order. Let us assume that function $M$, which is shown in Equation (\ref{eq:mapping}), is a mapping between indexes of corresponding elements in $X'$ and $X''$. By creating $W''$ from $W'$ using mapping $M$, it is possible to compute $f^t$ using Equation (\ref{eq:conv_sorted}). 
\begin{equation}
	\label{eq:neuron_computation}
	\small	f^t = \bigg(\sum_{i = 1}^N\sum_{j = 1}^{K}\sum_{k = 1}^{K} (W_{i, j, k} \times X_{i, j, k}^t) + b\bigg)^+
\end{equation}
\begin{equation}
	\begin{split}
		\label{eq:simp_cond}
		\small \forall~l \in \mathbb{N}_{NK^2}:\\
		\small if~~l = K \times (i~\times~&K + j) + k \Rightarrow W'_{l} = W_{i, j, k}~~and~~X'_{l} = X^t_{i, j, k} 
	\end{split}
\end{equation}
\begin{equation}
	\label{eq:simplified_conv}
	\small 	f^t = \bigg(\sum_{l = 1}^{NK^2} (W'_{l} \times X'_{l}) + b\bigg)^+
\end{equation}
\begin{equation}
	\label{eq:mapping}
	\small 	if~j = M(i) \Rightarrow X''_j = X'_i
\end{equation}
\begin{equation}
	\label{eq:conv_sorted}
	\small  f^t = \bigg(\sum_{l = 1}^{NK^2} (W''_{l} \times X''_{l}) + b\bigg)^+
\end{equation}

Since a considerable number of neurons are dormant in each layer, X is expected to be a sparse vector. Assuming a sparsity ratio of $\tau_t$ for $X$, one can compute $f^t$ using Equation~(\ref{eq:conv_sparse}). 
\begin{equation}
	\label{eq:conv_sparse}
	\small f^t = \bigg(\sum_{l = 1}^{(NK^2 - \tau_t NK^2)} (W''_{l} \times X''_{l}) + b\bigg)^+
\end{equation}

Neural networks are designed to be noise resilient. Hence, it is possible to approximate small elements in $X$ with zeros to increase the sparsity ratio from $\tau_t$ to $\tau'_t$. This can be performed using a cutoff threshold: $TH_t$.

The value of $\tau'_t$, which is an indicator of the number of neurons that did not fire for a given input, is called dormant neurons' density in the rest of the paper. The difference between $\tau_t$ (neurons which are actually dormant) and $\tau'_t - \tau_t$ (neurons that are considered dormant) is important. The former simply reflects the number of zeros in OFMs. However, the latter is the number of small elements that were approximated by zero without changing the classification results. The largest value of $TH_t$ for \textbf{each tile} of a given input can be found by performing a binary search on the range of possible values and observing the effect on the classification accuracy. 

For ease and efficiency of implementation, we compute a single sparsity ratio for each layer of a CNN (instead of each tile) whose upper bound can be calculated using Equation (\ref{eq:tau}) where $T$ is the total number of tiles in the IFMs of layer $l$. That is, for each layer of a CNN finding only one cutoff value ($TH_l$) suffices.

	\begin{equation}
		\label{eq:tau}
		\small \tau'_l \leq max\{\tau'_1, \tau'_2, \cdots, \tau'_T\}
	\end{equation}

The value of dormant neurons' density is expected to be smaller for shallow layers which mainly consist of CANs. However, it ought to be larger for deep layers which include a large number of CDNs.

\subsection{Case Study: Dormant Neurons' Density in Different Layers of GoogLeNet}
\begin{figure}
	\includegraphics[width=7cm]{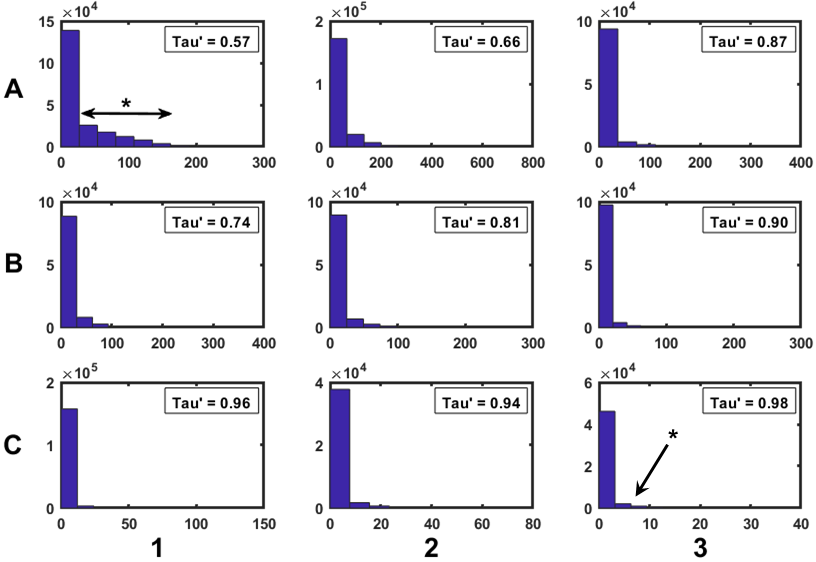}
	\caption{\small Histogram of OFMs of main layers of GoogLeNet for a sample input. By moving towards deeper layers, the value of $\tau'$, i.e., the number of neurons that have negligible impact on classification result, increases.  A1, A2, A3, B1, B2, B3, C1, C2, and C3 are outputs of layers: CL2, I3A, I4A, I4B, I4C, I4D, I4E, I5A, and I5B, respectively.}
	\label{Fig:ofm_hist}
\end{figure}

Figure \ref{Fig:ofm_hist} illustrates the histogram of OFM values for nine major layers of GoogLeNet. In the shallowest layer (A1), a considerable number of neurons are active (shown by an asterisk). This phenomenon was expected since most of the CANs which reside in the first layer collaborate to extract features that will be used by subsequent layers. As we move towards the deeper layers, the value of $\tau'$ gradually increases from 0.57 to 0.98. That is, in the very last Inception layer (C3), which mainly includes CDNs, 98\% of neurons do not contribute to classifying the given input. Comparing the asterisk in (A1) to the one in (C3) shows how activity patterns change as we approach the deepest layers of a CNN.

\begin{figure}
	\includegraphics[width=7cm]{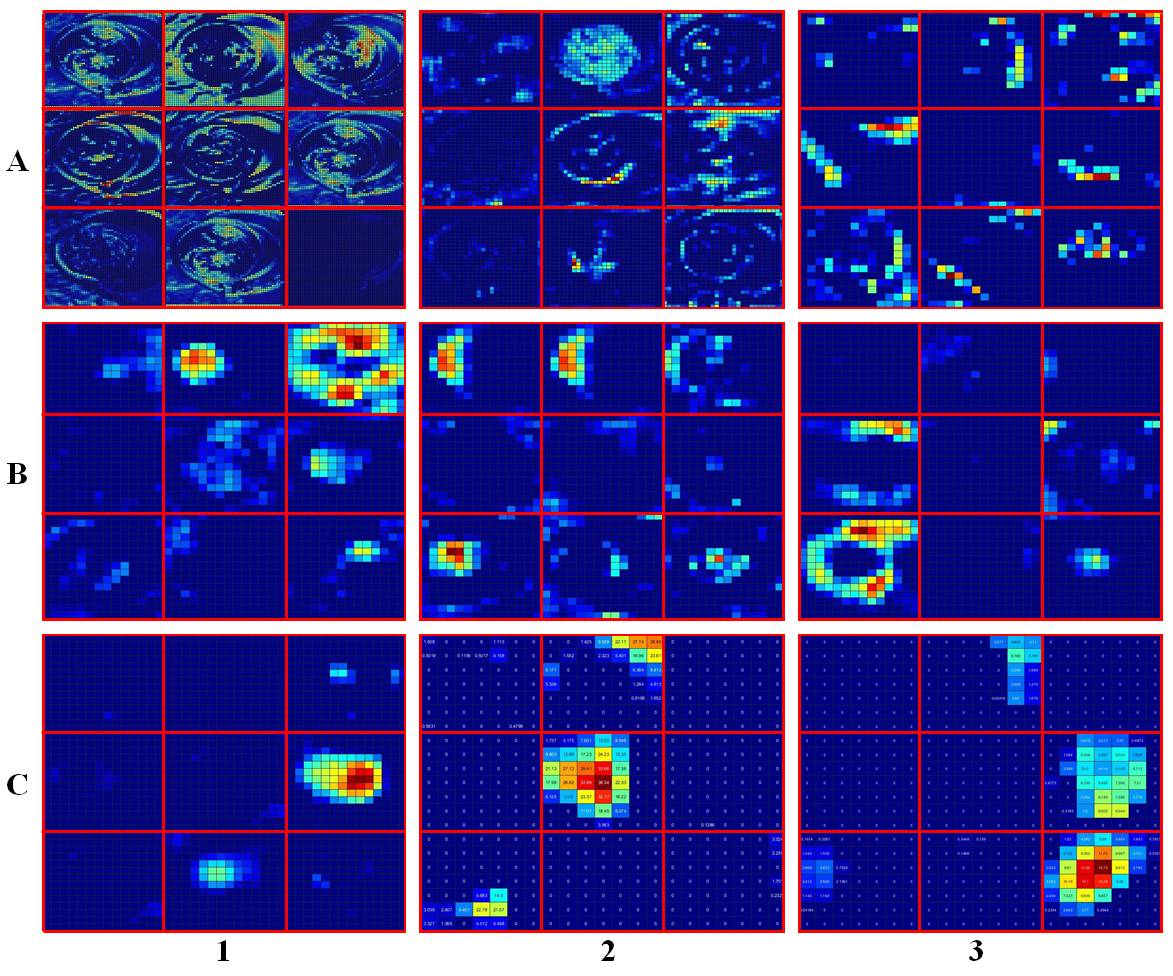}
	\caption{\small Nine random OFMs per each major layer of GoogLeNet. The order of layers is congruent with that of Figure~\ref{Fig:ofm_hist}. As we go deeper in the network, more neurons become dormant. However, those that fire have higher output values. This indicates that the CNN is forming a strong opinion about the label of the input.}
	\label{Fig:ofm_visual}
\end{figure}

We visualized nine random OFMs per each of the major layers of GoogLeNet and the results are depicted in Figure~\ref{Fig:ofm_visual} (the order of layers is congruent with that of Figure~\ref{Fig:ofm_hist}). There are two important patterns to observe. First, the variation of the dormant neurons' density from the very first layer (A1) to the very last one (C3), and how an increasing area of heatmaps gets cooler. Second, in the deep layers, some of the active neurons that fire have very high output values. Those seem to be mainly CDNs that have a high sensitivity for the class to which the given input belongs.

\section{CNN Distillation}
Running server-grade CNNs on resource-constrained IoT devices can be inefficient from various facets including:\\
\textbf{(1) Energy Consumption:} It is possible to efficiently accelerate CNNs on various IoT devices \cite{zhang2015optimizing,han2016eie,chen2017eyeriss,qiu2016going,motamedi2017cappuccino}. However, even after such an implementation, the required energy budget can be large for most battery-operated platforms. For example, Motamedi et al. showed that an efficient acceleration of GoogLeNet on mobile System-on-Chips (SoCs), requires 3.24J energy per inference~\cite{motamedi2017machine}.\\
\textbf{(2) Applicability in IoT Domains:} Many IoT applications demand fast and accurate understanding of a few key events in their surrounding environment. 
Recognizing unimportant events, which imposes extra stress on the energy source, is not the best use of our constrained resources in an IoT platform.
As an example, in developing an IoT application for pet monitoring, recognition of presence or absence of the animal in a camera feed can be sufficient. In such cases, we do not need the CNN to be able to distinguish the breed of the dog. Instead, it is expected to work reliably in terms of execution time, energy consumption, and heat dissipation. Nonetheless, using GoogLeNet for such tasks, executes all of its neurons, including those whose outputs are unimportant (e.x., breed detection) or not applicable (e.x., whale detector). 

On one hand, using very high-end CNNs for IoT-class applications can waste the available computing power. On the other hand, since training CANs benefits from data of all classes, retraining a new CNN with a limited number of training examples (only from those classes that belong to the scope of interest of a particular task) can lead to an inferior performance. In this section, we propose a principled methodology for identifying CANs and CDNs whose outputs are essential for understanding events that are in the interest of a particular IoT application. The detected neurons form a smaller CNN which is capable of fulfilling the required tasks while avoiding unnecessary computations.
\subsection{Neuron Removal}
\subsubsection{Complete Neuron Removal}
\begin{figure}
	\includegraphics[width=7cm]{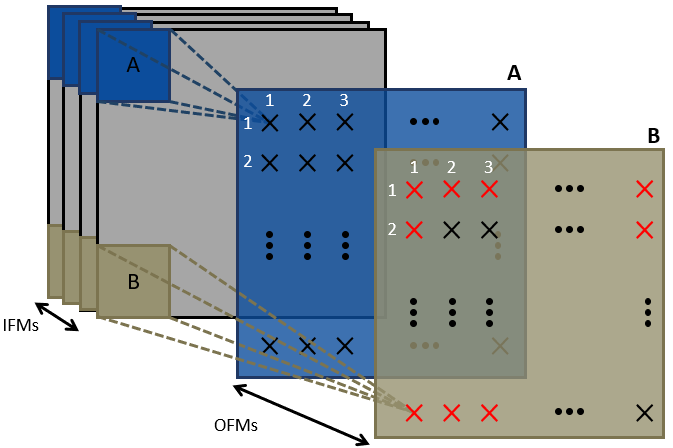}
	\caption{\small Dormant and active neurons. OFMs A and B in layer~\#$(l)$ contain the output of neurons A and B, respectively. A red cross in location $(i, j)$ of an OFM indicates that the corresponding neuron has fired in that particular position.} 
	\label{Fig:neuronExtraction}
\end{figure}

Let us assume that a neuron always remains dormant for a particular application. That is, it either never fires or, if it fires, the output value is negligible. This is illustrated in Figure~\ref{Fig:neuronExtraction}, where neuron A refuses to fire on all of the different spatial locations of IFMs. If this phenomenon deterministically repeats for a specific learning task, it is computationally beneficial to extract such a neuron. Therefore, neuron A in Figure~\ref{Fig:neuronExtraction} should be removed which leads to elimination of OFM A and saves all of the computations that should have been performed to create it. Neuron removal in a layer $l$ results in computation reduction in the same layer and computation reduction in the subsequent layer (since it decreases the number of OFMs in layer $l$. OFMs in layer $l$ are IFMs to layer $l + 1$). Equation (\ref{eq:savingL}) and Equation (\ref{eq:savingL1}) show the number of Multiply-Accumulation (MAC) operations which are skipped in layers $l$ and $l + 1$, respectively. In these equations, $W_{out}$, $H_{out}$, $N$, $K$, and $S$ stand for width of OFMs, height of OFMs, number of IFMs, kernel size, and convolution stride, respectively. Subscript $l$ emphasizes that values of parameters vary across different layers of the neural network.
\begin{equation}
	\label{eq:savingL}
	\small \# \text{skipped}~\text{MACs}_l = \frac{Wout_l \times Hout_l \times N_l \times K_l^2}{S_l^2}
\end{equation}
\begin{equation}
	\label{eq:savingL1}
	\small \# \text{skipped}~\text{MACs}_{l + 1} = \frac{Wout_{l + 1} \times Hout_{l + 1} \times M_{l + 1} \times K_{l + 1}^2}{S_{l + 1}^2}
\end{equation}
\subsubsection{Partial Neuron Removal}
As illustrated in Figure \ref{Fig:neuronExtraction}, there are cases where a neuron fires on some regions of IFMs and stays dormant on other regions (Neuron B). If a neuron remains dormant in a certain spatial location for a particular task, it is computationally efficient to skip calculating its value in that specific location.

The number of dormant neurons that can be removed varies based on the original CNN and the type of the IoT application that the CNN is being distilled for. This number also varies among different layers of a CNN. In what follows we continue our discussion by offering a principled approach for recognition and partial or complete removal of dormant neurons.
\subsection{Neuron Removal vs. Weight Pruning}
Weight pruning is concerned with manipulating neuron interconnections by approximating small weight values with zeros \cite{han2015deep}. Each pruning changes the topology of CNN architecture and attenuates the relationship of two neurons. Pruning increases the sparsity of weight matrix; however, that does not significantly drop the execution time since the current computing infrastructures are very inefficient when it comes to numerical calculation on non-uniform sparse matrices \cite{szegedy2015going,han2015deep,jouppi2017datacenter}.

\textbf{Fallacy}: If we realize $W_{i, j, k}$ in Equation (\ref{eq:neuron_computation}) equals zero during runtime, avoiding the computation of $W_{i, j, k} \times X^t_{i, j, k}$ drastically improves the power consumption.

The aforementioned statement is not quite accurate since under 45 nm CMOS technology, a 32-bit operation on average requires 1 pJ energy, while a 32-bit cache access demands 6 pJ and a 32-bit DRAM access demands 640 pJ. Given a hit rate of $h$, Equation~(\ref{eq:pruning}) compares the required energy for performing multiplication with the required energy for loading the operands. For instance, if cache's hit rate equals to 0.95, avoiding the multiplication improves the energy consumption by 2.58\%. That is, 97.42\% of the energy budget is dedicated to loading the weight value. An analogous analysis can be performed to show that the effect of such optimization is also limited on the execution time. In addition, using a branch to check the value of $W$ decreases the efficiency of speculative instruction execution which can further decrease the performance.
\begin{equation}
	\label{eq:pruning}
	\small  \text{energy}~\text{saving} = \frac{1}{1 + h \times 6 + (1 - h) \times 640} \times 100
\end{equation}

In contrast, neuron removal targets a group of weights which are semantically and structurally related (a neuron) and aims to eliminate all of them together when possible. A complete neuron removal requires no further overhead. That is, during runtime it is not required to check a flag to determine if a neuron had been removed. Hence, the energy consumption and execution time proportionally decrease by the sum of the values that Equation~(\ref{eq:savingL}) and Equation (\ref{eq:savingL1}) compute. For the partial neuron removals, however, a bitmap is required for bookkeeping purposes to determine if a neuron fires in a spatial location. For each location, if the bitmap's value is one, all of the computations of the corresponding neuron in that specific place has to be performed. Otherwise, all the instructions which includes $2NK ^2$ memory and $NK ^2$ MAC operations can be skipped. 

In addition, modern computing devices, especially GPUs, do not handle branching effectively \cite{shazeer2017outrageously}. The key difference between partial neuron removal, rather than pruning, is that removing a neuron impacts a large chunk of computations. Therefore, the outcome easily compensates the overhead of decision-making processes. In addition, complete neuron removal does not require branching.

After pruning a CNN, it must preserve its ability to classify different images. Hence, a pruned network can generate distinctive features that are used to discriminate between different classes. That is, in a pruned CNN there are kernels which extract features that are exclusively used for identification of images of a particular class. This is the sufficient condition for distillation to work. Hence it is possible to distill a pruned CNN or prune a distilled CNN.

It is worth noting that pruning removes weights which are not useful for \textbf{any} class, whereas distillation eliminates those neurons that are not beneficial for recognizing classes that belong to the \textbf{scope of interest of a particular IoT application}. Hence, distillation and pruning target different problems.
\subsection{CNN Distillation Algorithm}
\label{sec:CNN_Distillation}
\begin{figure*}
	\includegraphics[width=\linewidth]{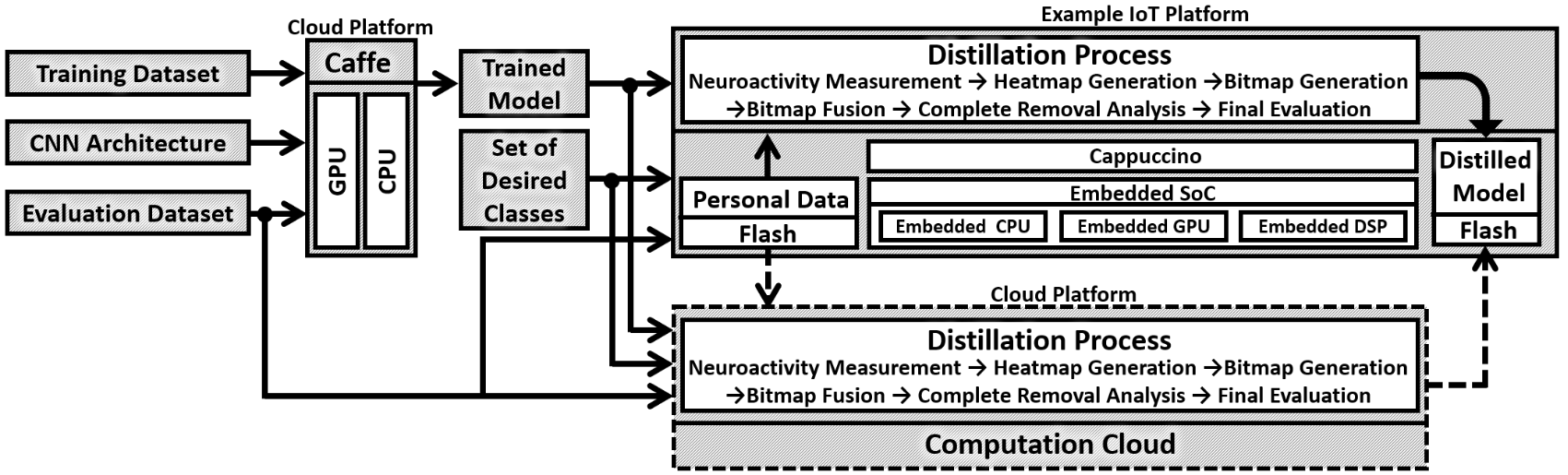}
	\caption{\small The distillation process takes a trained CNN and eliminates neurons which are responsible for understanding subjects that are out of an IoT application's scope of interest. The distillation process can be performed on a mobile device or a computation cloud.} 
	\label{Fig:blockdiagram}
\end{figure*}
\subsubsection{Neuroactivity Measurement} For each member of training dataset that belongs to the classes of set $B$ (from Equation~(\ref{eq:setAB})), the neural network has to be run in inference mode. The output of each layer, which is an indicator of neurons' reaction to each particular input, has to be preserved. We use notation $OFM_{b_i, l}^j$ to show the responses of neurons that reside in layer $l$ of the neural network to $j^{th}$ element of set $Train_{b_{i}}$. The set $Train_{b_{i}}$ includes all of the samples of the training section of ILSVRC dataset~\cite{ILSVRC15} that belong to the class $b_i$. The cardinality of this set is 1300 for most classes.\\
\subsubsection{Heatmap Generation} To detect all neurons which play a considerable role in recognizing members of a class, for every layer of the neural network, we need to compute the average neuron activities for each class $b_i$ using Equation (\ref{eq:avgact}). In this equation, $OFM_{b_i, l}$ indicates the average responses of neurons of layer $l$ to all training data of class $b_i$, and $OFM_{b_i, l}^j$ is computed in the last step.
\begin{equation}
		\label{eq:avgact}
		\small  OFM_{b_i, l} = \frac{\sum_{j = 1}^{|Train_{b_{i}}|}OFM_{b_i, l}^j}{|Train_{b_{i}}|}
\end{equation}
\subsubsection{Bitmap Generation}  The heatmap which is generated in the previous step ($OFM_{b_i, l}$) shows the average reaction of each neuron to elements of $Train_{b_{i}}$. This value is zero for most of CDNs and can be negligible for some CANs. Using a threshold $TH_l$, it is possible to sort out dormant neurons and save their location in a bitmap as it is shown in Equation~(\ref{eq:bitmap}) where, $M$ is the number of OFMs, $N$ is the number of IFMs, $Win$ is the width of IFMs, and $Hin$ is the height of IFMs. In addition, bitmap is a 3D data structure that keeps track of neurons whose computations can be skipped.
\begin{equation}
		\begin{split}
			\label{eq:bitmap}
			\forall~m \in \mathbb{N}_M, \forall~h \in \mathbb{N}_{Hin}, \forall~w \in \mathbb{N}_{Win}:\quad\quad\quad\quad\quad\quad\quad\quad \\bitmap_{b_i, l}[m][h][w] = 
			\begin{cases}
				\text{1} &OFM_{b_i, l}[m][h][w]\ge TH_l\\
				\text{0} &\text{otherwise}\\
			\end{cases} 	
		\end{split}
\end{equation}
\subsubsection{Bitmap Fusion} The bitmaps which are computed in Equation~(\ref{eq:bitmap}) are class specific. That is, applying $bitmap_{b_i, l}$ to a CNN eliminates all of its sections that do not contribute in classifying members of class $b_i$. In order to successfully fulfill our purpose, i.e., be able to classify members of set $B$, all of the corresponding bitmaps must be fused. This can be achieved using the Hadamard Product as it is shown in Equation~(\ref{eq:agr}) in which the ``$\sim$'' symbol shows an element-wise inversion.
\begin{equation}
		\label{eq:agr}
		bitmap_l =\sim (\sim bitmap_{b_1, l} \circ \sim bitmap_{b_2, l} \circ \cdots \circ \sim bitmap_{b_{|B|}, l})
\end{equation}
In Equation~(\ref{eq:agr}), $bitmap_l$ is a 3D data structure that shows neurons whose outputs are essential for understanding images that belong to classes of set $B$. Therefore, if $bitmap_l [m][h][w] = 0$, neuron $m$ of layer $l$ in location $(h, w)$ can be partially removed.
\subsubsection{Identifying Candidates for Complete Removal} The structure of bitmaps is analogous to that of OFMs. That is, each layer of it includes the outputs of a single neuron. Therefore, if all elements of layer $m$ in a bitmap equal to zero, the corresponding neuron ($m$) can be completely removed. In other words, neuron $m$ can be completely removed if the result of Equation~(\ref{eq:comrmv}) equals to zero.
\begin{equation}
	\label{eq:comrmv}
	\small \sum_{h = 1}^{Hout_l}\sum_{w = 1}^{Wout_l} bitmap_l [m][h][w]
\end{equation}

The distillation process is briefly illustrated in the block diagram of Figure \ref{Fig:blockdiagram}.
\subsection{The Required Time for Distilling a CNN}
The CNN distillation algorithm is designed with users' privacy protection in mind. Therefore, we developed it to be reasonably fast and energy efficient in order to be used on edge devices. This enables users to distill a CNN using their personal data without being forced to upload the data to an external computation cloud. In general, three important factors should be considered when it comes to algorithm design for IoT platforms:
\subsubsection{Energy Consumption} The distillation process has to be performed once. The results (i.e., neuron removal bitmaps) will be saved and can be used indefinitely. Given the offline nature of the process, it should be performed when the mobile platform is connected to an electrical grid. Hence, for this specific application, the impact of energy consumption is not as critical as other factors.
\subsubsection{Memory Footprint} During the distillation, a mobile platform has to keep $l$ different OFMs in the memory. Given the average OFM size of $28 \times 28 \times 522$, and average length of $60$ convolutional layers for deep CNNs, the required memory budget is 94 MB, which is affordable for most IoT devices that are qualified to host artificial intelligence algorithms.
\subsubsection{Execution Time} Motamedi et al. offered Cappuccino, which is a platform for inference software synthesis for mobile system-on-chips \cite{motamedi2017cappuccino}. Their approach requires 61.80 ms for each inference using AlexNet on a mediocre mobile platform (Google Nexus 6P). Since our distillation algorithm only requires the inference information, running it for a dataset of 10K private images on the aforementioned device requires 10.3 minutes of computation. 

When privacy is not an issue, the neural network distillation can be performed on a powerful workstation using platforms such as Caffe \cite{jia2014caffe} or TensorFlow \cite{abadi2016tensorflow}. Subsequently, the computed bitmaps which include guidance for neuron removal will be transferred to the target IoT device. 
\section{Results and Discussions}
\subsection{Experimental Setup}
In this section, we use GoogLeNet as a deep CNN to perform different experiments. GoogLeNet, which is trained on ILSVRC dataset~\cite{ILSVRC15}, has the state of the art performance in classification and is currently being used by Google in their computation cloud~\cite{jouppi2017datacenter}. In our experiments, we use the ILSVRC 2012 train dataset for network distillation process and take advantage of the validation part of the same dataset to measure the effect of neural network distillation on the classification accuracy. The ILSVRC dataset contains images of both natural and human objects with labels that indicate the presence or absence of an object in a scene. The dataset serves as one of the standard benchmarks for which novel CNN architectures can evaluate their performance.

Currently, there are two major approaches for measuring the classification accuracy: top-1 accuracy rate and top-5 accuracy rate~\cite{ILSVRC15}. The former compares the ground truth against the prediction with the highest probability; however, the latter compares the ground truth against the top 5 predictions. The classification result is accepted if the ground truth exists among the top five predictions. It is difficult to achieve a high top-1 accuracy. Nonetheless, in most real-world situations, a CNN is expected to have one correct prediction. Hence, in all of our experiments, we use the top-1 accuracy rate. The classification accuracy is measured on the preserved classes.

Even though techniques such as neural network ensembling and excessive cropping can be used to improve the classification accuracy in competitions, in resource-constrained IoT devices performing such computations is infeasible or inefficient. Therefore, in all of our experiments we use a single crop per input frame and only query a single CNN. Distilling a CNN for set $B$ (Equation~\ref{eq:setAB}) requires $\sum_{i=1}^{\beta}|Train_{b_{i}}|$ training examples where $Train_{b_{i}}$ is the set of training data that belongs to class $b_i$. Since the cardinality of $Train_{b_{i}}$ is approximately identical for different classes, the required data for distillation is $\beta/\alpha$ of the required data for training the CNN, where $\alpha$ is the number of all classes available in the dataset.
\subsection{Case Study: Emergency Vehicle Recognition}
\subsubsection{Application-Specific CNN Distillation}
GoogLeNet is trained to recognize classes that include emergency vehicles as well as many other classes which are not in the scope of interest of this particular task. The goal is to distill this CNN to remove those of its neurons that do not contribute to recognizing emergency vehicles.

\begin{figure}
	\includegraphics[width=7cm]{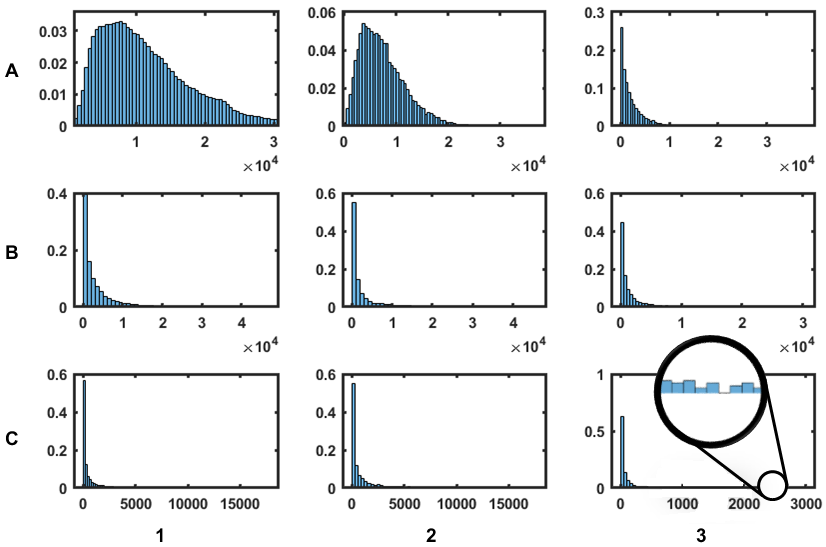}
	\caption{\small Normalized histogram of heatmap values for the most important layers of GoogLeNet. The neural network is being distilled to mainly focus on recognizing emergency vehicles. The order of CNN layers is congruent with that of Figure \ref{Fig:ofm_hist}.}
	\label{Fig:hist3}
\end{figure}

If desired, it is always possible to lose some accuracy for achieving higher neuron removal rates. However, all distillation examples that are mentioned in this paper are performed without sacrificing accuracy. We distilled the neural network according to the algorithm which is described in subsection \ref{sec:CNN_Distillation} to compute the heatmaps. The normalized histograms of heatmap values for the major layers of GoogLeNet are shown in Figure~\ref{Fig:hist3}. The order of CNN layers is congruent with the one in Figure \ref{Fig:ofm_hist}. As we expected, in the second convolutional layer, where most CANs that are structurally related to the input reside, the number of dormant neurons is very small. However, as we move towards the end of the neural network, this number starts to increase.

As it is highlighted in the last Inception layer (C3 in Figure~\ref{Fig:hist3}), a relatively small number of neurons have very large output values in deep layers. Even though those neurons are small in count, they are important in determining the outcome of the classification.
\subsubsection{Experimental Results}
\label{sec:Experimental Results}
Since a complete neuron removal in layer $l$ eliminates an entire OFM, the effect of such neuron removal on the required computation is identical to the impact of $Wout_l \times Hout_l$ partial neuron removal. Therefore, in the interest of simplicity, we count each complete neural removal as $Wout_l \times Hout_l$ partial neuron removal for the rest of the paper. Let us use Compression Ratio (CR) variable, which is defined in Equation~(\ref{eq:cr}), to demonstrate the amount of computations that could be skipped in distilling a CNN for a particular task. In Equation (\ref{eq:cr}), $pr$ and $cr$ are the number of neurons that are partially removed and the number of neurons that are completely removed, respectively.
\begin{equation}
	\label{eq:cr}
	\small \text{CR}_l = \frac{W_{out_{l}} \times H_{out_{l}} \times M_l}{W_{out_{l}} \times H_{out_{l}} \times M_l - (cr_l \times (Wout_l \times Hout_l) + pr_l)} \times 100
\end{equation}

It is worth noting that a decision on partial or complete removal of a neuron impacts a large chunk of computations. That is, for each and every single weight of that specific neuron, we skip the following operations: first, loading the weight data, second, computing the MAC result, and third, writing the output back.

Table \ref{tab:cr3} shows the compression ratio of each major layer of GoogLeNet for distilling it to another neural network which is capable of recognizing emergency vehicles only. 
\begin{table*}
	\caption{\small Compression ratios of different layers of GoogLeNet for distilling it to a smaller CNN that is capable of recognizing emergency vehicles. Numbers that are reported for each Inception layer are averaged over all convolutional layers that exist in that layer.}
	\label{tab:cr3}
	\scalebox{0.8}{
			\begin{tabular}{cccccccccccccc}
				\toprule[.07cm]
				& I3A & I3B & I4A & I4B & I4C & I4D & I4E & I5A & I5B & Total MFLOPs \cite{szegedy2015going} & MFLOPs Skipped & Speedup & Accuracy Loss \\ \midrule
				&  1.05X  &  1.65X  &  1.71X  &  1.68X  &  2.05X  &  2.86X  &  3.49X  &  4.26X  &  4.71X  &                 \textbf{1107}                 &      \textbf{539}          &    1.95X     & \textbf{0.0\%} \\ \bottomrule[.07cm]
			\end{tabular}
	}
\end{table*}

There are four notable points regarding the compression rates that are presented in Table \ref{tab:cr3}:\\
(1) Starting from Inception layer 3B, the compression rates become considerable. A compression rate of 1.65X in that layer indicates that the distillation process eliminates 40\% of unnecessary computations. This number increases semi-monotonically as we move towards deeper layer of the neural network. In the very last Inception layer, a compression rate of 4.71X means 78.78\% of the unnecessary computations of that layer are removed.\\
(2) In Inception layer 3A, the compression rate is negligible. This is the phenomenon that we expected. That is, CANs which have a higher density in shallow layers cannot be removed.\\
(3) The distillation is performed with classification accuracy preservation mindset. Therefore, the classification accuracy of the CNN on the classes that include emergency vehicles is not reduced after distillation. However, one can achieve higher compression rates, when small reduction of accuracy is tolerable. \\
(4) The distillation process, which only needs to be performed once per application, takes 39 minutes on a Qualcomm Snapdragon 810 SoC when the algorithm is accelerated using Cappuccino \cite{motamedi2017cappuccino}. We also implemented the algorithm using Caffe \cite{jia2014caffe}. The Caffe-based implementation finished the distillation process in 30.09 seconds using a Tesla K40 GPU.
\subsection{GoogLeNet Distillation for Random Classes}
\label{sec:GoogLeNet Distillation for Random Classes}
Classes that contain emergency vehicles are semantically related. That is, all emergency vehicles are some sort of cars and share a number of feature-sets, such as wheels and headlights, that a CNN uses to recognize vehicles. This leads to a higher compression rate since classes that belong to our scope of interest do not demand a high variety of CDNs. To study the impact of distillation on subsets which includes classes with no particular semantical relationship, we used the distillation algorithm offered in Section \ref{sec:CNN_Distillation} to distill GoogLeNet for three different subsets of ILSVRC datasets whose classes are chosen randomly. The first, second, and third subset contains 5, 10, and 15 random classes, respectively.
\begin{table*}
	\caption{\small Compression of major layers of GoogLeNet for distilling it to smaller CNNs tailored to recognize members of three sets that we mentioned in Subsection \ref{sec:GoogLeNet Distillation for Random Classes}. Numbers that are reported for each Inception Layer are averaged over all convolutional layers that it includes.}
	\label{tab:crall}
	\scalebox{0.8}{
			\begin{tabular}{cccccccccccccc}
				\toprule[.07cm]
				& I3A & I3B & I4A & I4B & I4C & I4D & I4E & I5A & I5B & Total MFLOPs \cite{szegedy2015going} & MFLOPs Skipped & Speedup & Accuracy Loss \\ \toprule[.07cm]
				Subset 3 &  1.01X  &  1.43X  &  1.44X  &  1.13X  &  1.50X  &  1.80X  &  1.46X  &  1.38X  &  1.98X  &            \textbf{1107}             &       \textbf{315}       &  1.4X   &       \textbf{0.00\%}      \\
				Subset 2 &  1.02X  &  1.43X  &  1.53X  &  1.30X  &  1.73X  &  2.28X  &  1.90X  &  1.47X  &  2.02X  &            \textbf{1107}             &       \textbf{382}       &  1.5X   &       \textbf{0.20\%}      \\
				Subset 1 &  1.03X  &  1.46X  &  2.09X  &  1.34X  &  1.86X  &  2.83X  &  3.25X  &  2.24X  &  4.35X  &            \textbf{1107}             &       \textbf{485}       &  1.8X   &       \textbf{0.13\%}      \\ \bottomrule[.07cm]
			\end{tabular}
	}
\end{table*}

The results of GoogLeNet distillation for these classes are shown in Table \ref{tab:crall}. Increasing the diversity of elements of a subset for which a CNN is being distilled can have an adverse effect on the achievable compression ratio. Augmenting the cardinality of the subset of interest and handpicking the elements of that subset to have minimal semantical relationship with each other are both effective in increasing the diversity. In the experiments whose results are shown in Table \ref{tab:crall}, both approaches are used. Since for a subset with a high diversity a wide variety of neurons have to contribute to make the classification process successful, it is expected to observe smaller compression ratio as diversity increases. In an extreme case, when the size of the subset equals the number of classes that the CNN is originally trained on, we expect the compression rate to become one.
\subsection{Distillation on CIFAR and MNIST Datasets}
To further inspect the effectiveness of the proposed approach, we distilled two other CNNs: The first CNN is the TensorFlow~\cite{abadi2016tensorflow} reference design for classifying CIFAR-10~\cite{krizhevsky2014cifar} dataset and the second one is the reference design for classifying MNIST~\cite{lecun1998mnist}. The CIFAR-10 dataset is a collection of color images which contains 50000 training and 10000 test samples. The MNIST dataset includes handwritten images, and has a training set of 60000 samples, and a test set of 10000 samples.

In each experiment, the number of target classes is decreased, the CNN is distilled, and the effect on the parameter reduction is studied. In a CNN with 10 classes, one can hope that at most 10\% of all parameters are exclusively used for understanding each class. Therefore, in an ideal setting, removing a class and distilling the CNN can at most eliminate 10\% of the parameters. We use the ideal case as a reference to gauge the effectiveness of the distillation. Table~\ref{tab:mnist-cifar} presents the distillation results on the aforementioned CNNs. The proposed algorithm can achieve near-linear parameter reduction specifically when a smaller number of classes are removed. In all experiments, the loss in classification accuracy is restricted to be at most 1\%. The distillation is performed on the training subset of each dataset (50000 samples for CIFAR-10, and 60000 samples for MNIST). Subsequently, the distilled CNNs are tested on the evaluation subset of the datasets.
\begin{table}
	\caption{\small Distillation results on the TensorFlow reference designs for MNIST and CIFAR-10 datasets. In all experiments, the accuracy loss is projected to be less than 1.0\%. The fine-tuning column shows the results that a fine-tuning-based approach would yield.}
	\label{tab:mnist-cifar}
	\scalebox{0.75}{
		\begin{tabular}{cccccccc}
				\toprule[.07cm]
				\multirow{2}{*}{\rotatebox{90}{\bf \small CNN}}& \multirow{2}{*}{\makecell{\bf \small Removed \\\bf \small Classes (\%)}} & \multirow{2}{*}{\makecell{\bf \small Total\\\bf \small KFLOPS}} & \multirow{2}{*}{\makecell{\bf \small Skipped\\\bf \small KFLOPS}} & \multicolumn{3}{c}{\bf \small Parameter Removal (\%)} & \multirow{2}{*}{{\makecell{\bf \small Accuracy \\ \bf \small Loss}}}\\\cmidrule{5-7}
				&&&&\bf \small Achieved &\bf \small Fine-tuning&\bf \small Ideal &
				\\ \toprule[.07cm]
				\multirow{11}{*}{\rotatebox{90}{TF-MNIST}}
				&10&18014&1801&10&0.008&10&0.02\%\\\cmidrule{2-8}
				&20&18014&3602&20&0.016&20&0.05\%\\\cmidrule{2-8}
				&30&18014&5403&30&0.024&30&0.06\%\\\cmidrule{2-8}	
				&40&18014&7204&40&0.031&40&0.27\%\\\cmidrule{2-8}					
				&50&18014&9555&49&0.039&50&0.97\%\\\cmidrule{2-8}
				&60&18014&9555&49&0.047&60&0.99\%\\\cmidrule{2-8}
				&70&18014&10808&57&0.055&70&0.98\%\\\cmidrule{2-8}
				&80&18014&13737&71&0.063&80&0.93\%\\ \bottomrule[.07cm]			
				\multirow{11}{*}{\rotatebox{90}{TF-CIFAR}}
				&10&25112&2511&10&0.022&10&0.01\%\\\cmidrule{2-8}
				&20&25112&5022&20&0.045&20&0.01\%\\\cmidrule{2-8}
				&30&25112&7534&30&0.064&30&0.0\%\\\cmidrule{2-8}
				&40&25112&10045&40&0.09&40&0.01\%\\\cmidrule{2-8}
				&50&25112&12556&50&0.112&50&0.06\%\\\cmidrule{2-8}
				&60&25112&15067&60&0.135&60&0.13\%\\\cmidrule{2-8}
				&70&25112&17277&68&0.157&70&0.48\%\\\cmidrule{2-8}
				&80&25112&19525&76&0.179&80&0.93\%\\\bottomrule[.07cm]
		\end{tabular}
	}
\end{table}
\begin{table*}
	\caption{\small Achieved experimental speedups in running the distilled version of TF-MNIST on Google Nexus 6P. Each experiment has been repeated 10 times and the results are reported in the form of $a \pm s$, where $a$ is the average, and $s$ is the standard deviation.}
	\label{tab:mnist-Nexus}
	\scalebox{0.9}{
		\begin{tabular}{cccccccccc}
				\toprule[.07cm]
				& \multirow{2}{*}{Baseline} &  \multicolumn{8}{c}{Removed Classes (\%)}\\\cmidrule{3-10}
				&&10&20&30&40&50&60&70&80\\ \toprule[.07cm]
				Execution Time (ms)&$213 \pm 5$&$199 \pm 5$  &$180 \pm 4$  &$152 \pm 4$  &$132 \pm 1$&$112 \pm 5$  &$87 \pm 2$  &$69 \pm 3$ &$50 \pm 2$\\\midrule
				Speedup            &N/A &1.07X&1.18X&1.40X&1.61X&1.90X&2.45X&3.09X&4.26X\\\midrule
				Ideal Speedup      &N/A &1.11X&1.25X&1.43X&1.67X&2.00X&2.50X&3.33X&5.00X\\\bottomrule[.07cm]
		\end{tabular}
	}
\end{table*}
\section{Related Work}
Deep CNNs, which involve many millions of parameters and billions of operations, are very compute-intensive. Such a huge computational demand is out of the budget for most of contemporary IoT platforms. Therefore, currently the mainstream approach for using artificial intelligence algorithms on such platforms is cloud-based computation. 

To address this issue, the research community has put forth a number of ways to improve the computational complexity of such algorithms on IoT platforms. In what follows we briefly review those approaches and indicate where our proposal stands.\\
(1) \textbf{Bit-compression and Quantization:} It has been shown that inference using CNNs do not necessarily require 32 bits floating point operations \cite{han2015deep,gysel2016hardware,hubara2016quantized,andri2016yodann}. The process can be performed using 16 bits or less fixed point arithmetic. This is particularly very beneficial for FPGA-based implementation of CNNs where we have the luxury of tailoring the architecture to the algorithm's requirements.\\
(2) \textbf{Compact Model Design:} Instead of trying to compress cumbersome CNNs, it is more effective to  develop a compact model in the first place. The expand-reduce model which is used in GoogLeNet \cite{szegedy2015going} and SqueezeNet \cite{iandola2016squeezenet} is a notable example. The idea behind SqueezeNet \cite{iandola2016squeezenet} is used to develop CNNs that address other vision applications such as point cloud segmentation \cite{wu2017squeezeseg}. MobileNet~\cite{howard2017mobilenets} and SqueezeNext~\cite{gholami2018squeezenext} are two other CNNs which are designed with a constrained parameter budget mindset. MobileNet~\cite{howard2017mobilenets} further improves AlexNet~\cite{krizhevsky2012imagenet} in terms of parameter budget while keeping the same accuracy. SqueezeNext~\cite{gholami2018squeezenext} further improves MobileNet~\cite{howard2017mobilenets} by decreasing the parameter budget by 13\%. Some of the proposed compact CNNs use residual layers such as DenseNet~\cite{huang2017densely} and CondenseNet~\cite{huang2017condensenet}. Huang et al. further improved DenseNet~\cite{huang2017densely} by introducing anytime classification~\cite{huang2017multi}. \\
(3) \textbf{FPGA-based Acceleration:} There are numerous proposals for FPGA-based CNN inference \cite{zhang2015optimizing,qiu2016going,chen2017eyeriss,ovtcharov2015accelerating}. However, there are two major issues constraining widespread usage of FPGA-based CNN acceleration. First, the time-consuming nature of hardware acceleration which increases the engineering costs. Second, the fact that not all IoT devices are equipped with an FPGA fabric.\\
(4) \textbf{Mobile SoC-based Acceleration:} Mobile SoC-based platforms are projected to be a major player in the emerging IoT landscape. Hence, an increasing number of solutions for accelerating CNNs on such platforms has been proposed~\cite{latifi2016cnndroid,alzantot2017rstensorflow,motamedi2017cappuccino,motamedi2017machine}.

Our approach is different from the previous work in that we use an application-oriented approach to remove computations whose results will never be used. The CNN distillation algorithm that we proposed in this paper is orthogonal and complementary to the aforementioned approaches. We tested the proposed algorithm on a CNN with an expand-reduce architecture (GoogLeNet \cite{szegedy2015going}) and used a mobile SoC-based inference software synthesizer to accelerate it (Cappuccino \cite{motamedi2017cappuccino}). The underlying idea in distillation is drastically different from other approaches. A working CNN must possess kernels that generate distinctive features which make it possible to discriminate between different classes. Distill-Net takes advantage of this characteristic to detect those neurons that are not beneficial for recognizing classes that belong to the scope of interest of a particular IoT application. To the best of our knowledge, an approach for CNN mission reassignment is proposed in this article for the first time.
\section{Conclusion}
In this paper, we proposed a systematic approach for recognition and elimination of those sections of a neural network which will never be used in a particular application. First, we studied how different neurons of a CNN react to a stimulus and discussed how class-agnostics neurons and class-dependent neurons contribute to understanding an input. Subsequently, we offered a distillation algorithm that can recognize those neurons which are dormant for all instances that belong to a particular set of classes. Finally, we offered an approach for removing such neurons to avoid performing computations that are not necessary for a specific application. Our experimental results confirms that the approach strikes a favorable balance between classification accuracy and inference efficiency.
\begin{acks}
	We would like to thank NVIDIA for donating GPUs that made this research possible.
\end{acks}
\bibliographystyle{ACM-Reference-Format}
\bibliography{bibliography}


\begin{thebibliography}{35}


\ifx \showCODEN    \undefined \def \showCODEN     #1{\unskip}     \fi
\ifx \showDOI      \undefined \def \showDOI       #1{#1}\fi
\ifx \showISBNx    \undefined \def \showISBNx     #1{\unskip}     \fi
\ifx \showISBNxiii \undefined \def \showISBNxiii  #1{\unskip}     \fi
\ifx \showISSN     \undefined \def \showISSN      #1{\unskip}     \fi
\ifx \showLCCN     \undefined \def \showLCCN      #1{\unskip}     \fi
\ifx \shownote     \undefined \def \shownote      #1{#1}          \fi
\ifx \showarticletitle \undefined \def \showarticletitle #1{#1}   \fi
\ifx \showURL      \undefined \def \showURL       {\relax}        \fi
\providecommand\bibfield[2]{#2}
\providecommand\bibinfo[2]{#2}
\providecommand\natexlab[1]{#1}
\providecommand\showeprint[2][]{arXiv:#2}

\bibitem[\protect\citeauthoryear{Abadi, Agarwal, Barham, Brevdo, Chen, Citro,
  Corrado, Davis, Dean, Devin, Ghemawat, Goodfellow, Harp, Irving, Isard, Jia,
  Jozefowicz, Kaiser, Kudlur, Levenberg, Man\'{e}, Monga, Moore, Murray, Olah,
  Schuster, Shlens, Steiner, Sutskever, Talwar, Tucker, Vanhoucke, Vasudevan,
  Vi\'{e}gas, Vinyals, Warden, Wattenberg, Wicke, Yu, and Zheng}{Abadi
  et~al\mbox{.}}{2015}]%
        {abadi2016tensorflow}
\bibfield{author}{\bibinfo{person}{Mart\'{\i}n Abadi}, \bibinfo{person}{Ashish
  Agarwal}, \bibinfo{person}{Paul Barham}, \bibinfo{person}{Eugene Brevdo},
  \bibinfo{person}{Zhifeng Chen}, \bibinfo{person}{Craig Citro},
  \bibinfo{person}{Greg~S. Corrado}, \bibinfo{person}{Andy Davis},
  \bibinfo{person}{Jeffrey Dean}, \bibinfo{person}{Matthieu Devin},
  \bibinfo{person}{Sanjay Ghemawat}, \bibinfo{person}{Ian Goodfellow},
  \bibinfo{person}{Andrew Harp}, \bibinfo{person}{Geoffrey Irving},
  \bibinfo{person}{Michael Isard}, \bibinfo{person}{Yangqing Jia},
  \bibinfo{person}{Rafal Jozefowicz}, \bibinfo{person}{Lukasz Kaiser},
  \bibinfo{person}{Manjunath Kudlur}, \bibinfo{person}{Josh Levenberg},
  \bibinfo{person}{Dandelion Man\'{e}}, \bibinfo{person}{Rajat Monga},
  \bibinfo{person}{Sherry Moore}, \bibinfo{person}{Derek Murray},
  \bibinfo{person}{Chris Olah}, \bibinfo{person}{Mike Schuster},
  \bibinfo{person}{Jonathon Shlens}, \bibinfo{person}{Benoit Steiner},
  \bibinfo{person}{Ilya Sutskever}, \bibinfo{person}{Kunal Talwar},
  \bibinfo{person}{Paul Tucker}, \bibinfo{person}{Vincent Vanhoucke},
  \bibinfo{person}{Vijay Vasudevan}, \bibinfo{person}{Fernanda Vi\'{e}gas},
  \bibinfo{person}{Oriol Vinyals}, \bibinfo{person}{Pete Warden},
  \bibinfo{person}{Martin Wattenberg}, \bibinfo{person}{Martin Wicke},
  \bibinfo{person}{Yuan Yu}, {and} \bibinfo{person}{Xiaoqiang Zheng}.}
  \bibinfo{year}{2015}\natexlab{}.
\newblock \bibinfo{title}{{TensorFlow}: Large-Scale Machine Learning on
  Heterogeneous Systems}.
\newblock
\newblock
\urldef\tempurl%
\url{https://www.tensorflow.org/}
\showURL{%
\tempurl}
\newblock
\shownote{Software available from tensorflow.org.}


\bibitem[\protect\citeauthoryear{Alzantot, Wang, Ren, and Srivastava}{Alzantot
  et~al\mbox{.}}{2017}]%
        {alzantot2017rstensorflow}
\bibfield{author}{\bibinfo{person}{Moustafa Alzantot}, \bibinfo{person}{Yingnan
  Wang}, \bibinfo{person}{Zhengshuang Ren}, {and} \bibinfo{person}{Mani~B
  Srivastava}.} \bibinfo{year}{2017}\natexlab{}.
\newblock \showarticletitle{Rstensorflow: Gpu enabled tensorflow for deep
  learning on commodity android devices}. In
  \bibinfo{booktitle}{\emph{Proceedings of the 1st International Workshop on
  Deep Learning for Mobile Systems and Applications}}. ACM,
  \bibinfo{pages}{7--12}.
\newblock


\bibitem[\protect\citeauthoryear{Andri, Cavigelli, Rossi, and Benini}{Andri
  et~al\mbox{.}}{2016}]%
        {andri2016yodann}
\bibfield{author}{\bibinfo{person}{Renzo Andri}, \bibinfo{person}{Lukas
  Cavigelli}, \bibinfo{person}{Davide Rossi}, {and} \bibinfo{person}{Luca
  Benini}.} \bibinfo{year}{2016}\natexlab{}.
\newblock \showarticletitle{YodaNN: An ultra-low power convolutional neural
  network accelerator based on binary weights}. In
  \bibinfo{booktitle}{\emph{VLSI (ISVLSI), 2016 IEEE Computer Society Annual
  Symposium on}}. IEEE, \bibinfo{pages}{236--241}.
\newblock


\bibitem[\protect\citeauthoryear{Atzori, Iera, and Morabito}{Atzori
  et~al\mbox{.}}{2010}]%
        {atzori2010internet}
\bibfield{author}{\bibinfo{person}{Luigi Atzori}, \bibinfo{person}{Antonio
  Iera}, {and} \bibinfo{person}{Giacomo Morabito}.}
  \bibinfo{year}{2010}\natexlab{}.
\newblock \showarticletitle{The internet of things: A survey}.
\newblock \bibinfo{journal}{\emph{Computer networks}} \bibinfo{volume}{54},
  \bibinfo{number}{15} (\bibinfo{year}{2010}), \bibinfo{pages}{2787--2805}.
\newblock


\bibitem[\protect\citeauthoryear{Bandyopadhyay and Sen}{Bandyopadhyay and
  Sen}{2011}]%
        {bandyopadhyay2011internet}
\bibfield{author}{\bibinfo{person}{Debasis Bandyopadhyay} {and}
  \bibinfo{person}{Jaydip Sen}.} \bibinfo{year}{2011}\natexlab{}.
\newblock \showarticletitle{Internet of things: Applications and challenges in
  technology and standardization}.
\newblock \bibinfo{journal}{\emph{Wireless Personal Communications}}
  \bibinfo{volume}{58}, \bibinfo{number}{1} (\bibinfo{year}{2011}),
  \bibinfo{pages}{49--69}.
\newblock


\bibitem[\protect\citeauthoryear{Bhagyesh~Vikani}{Bhagyesh~Vikani}{2017}]%
        {tf_cnnvis}
\bibfield{author}{\bibinfo{person}{Falak~Shah Bhagyesh~Vikani}.}
  \bibinfo{year}{2017}\natexlab{}.
\newblock \bibinfo{title}{CNN Visualization}.
\newblock \bibinfo{howpublished}{\url{https://github.com/InFoCusp/tf_cnnvis/}}.
\newblock


\bibitem[\protect\citeauthoryear{Chen, Choi, Yu, Han, and Chandraker}{Chen
  et~al\mbox{.}}{2017a}]%
        {chen2017learning}
\bibfield{author}{\bibinfo{person}{Guobin Chen}, \bibinfo{person}{Wongun Choi},
  \bibinfo{person}{Xiang Yu}, \bibinfo{person}{Tony Han}, {and}
  \bibinfo{person}{Manmohan Chandraker}.} \bibinfo{year}{2017}\natexlab{a}.
\newblock \showarticletitle{Learning Efficient Object Detection Models with
  Knowledge Distillation}. In \bibinfo{booktitle}{\emph{Advances in Neural
  Information Processing Systems}}. \bibinfo{pages}{742--751}.
\newblock


\bibitem[\protect\citeauthoryear{Chen, Krishna, Emer, and Sze}{Chen
  et~al\mbox{.}}{2017b}]%
        {chen2017eyeriss}
\bibfield{author}{\bibinfo{person}{Yu-Hsin Chen}, \bibinfo{person}{Tushar
  Krishna}, \bibinfo{person}{Joel~S Emer}, {and} \bibinfo{person}{Vivienne
  Sze}.} \bibinfo{year}{2017}\natexlab{b}.
\newblock \showarticletitle{Eyeriss: An energy-efficient reconfigurable
  accelerator for deep convolutional neural networks}.
\newblock \bibinfo{journal}{\emph{IEEE Journal of Solid-State Circuits}}
  \bibinfo{volume}{52}, \bibinfo{number}{1} (\bibinfo{year}{2017}),
  \bibinfo{pages}{127--138}.
\newblock


\bibitem[\protect\citeauthoryear{Gholami, Kwon, Wu, Tai, Yue, Jin, Zhao, and
  Keutzer}{Gholami et~al\mbox{.}}{2018}]%
        {gholami2018squeezenext}
\bibfield{author}{\bibinfo{person}{Amir Gholami}, \bibinfo{person}{Kiseok
  Kwon}, \bibinfo{person}{Bichen Wu}, \bibinfo{person}{Zizheng Tai},
  \bibinfo{person}{Xiangyu Yue}, \bibinfo{person}{Peter Jin},
  \bibinfo{person}{Sicheng Zhao}, {and} \bibinfo{person}{Kurt Keutzer}.}
  \bibinfo{year}{2018}\natexlab{}.
\newblock \showarticletitle{SqueezeNext: Hardware-Aware Neural Network Design}.
\newblock \bibinfo{journal}{\emph{arXiv preprint arXiv:1803.10615}}
  (\bibinfo{year}{2018}).
\newblock


\bibitem[\protect\citeauthoryear{Gysel, Motamedi, and Ghiasi}{Gysel
  et~al\mbox{.}}{2016}]%
        {gysel2016hardware}
\bibfield{author}{\bibinfo{person}{Philipp Gysel}, \bibinfo{person}{Mohammad
  Motamedi}, {and} \bibinfo{person}{Soheil Ghiasi}.}
  \bibinfo{year}{2016}\natexlab{}.
\newblock \showarticletitle{Hardware-oriented approximation of convolutional
  neural networks}.
\newblock \bibinfo{journal}{\emph{arXiv preprint arXiv:1604.03168}}
  (\bibinfo{year}{2016}).
\newblock


\bibitem[\protect\citeauthoryear{Han, Liu, Mao, Pu, Pedram, Horowitz, and
  Dally}{Han et~al\mbox{.}}{2016}]%
        {han2016eie}
\bibfield{author}{\bibinfo{person}{Song Han}, \bibinfo{person}{Xingyu Liu},
  \bibinfo{person}{Huizi Mao}, \bibinfo{person}{Jing Pu},
  \bibinfo{person}{Ardavan Pedram}, \bibinfo{person}{Mark~A Horowitz}, {and}
  \bibinfo{person}{William~J Dally}.} \bibinfo{year}{2016}\natexlab{}.
\newblock \showarticletitle{EIE: efficient inference engine on compressed deep
  neural network}. In \bibinfo{booktitle}{\emph{Computer Architecture (ISCA),
  2016 ACM/IEEE 43rd Annual International Symposium on}}. IEEE,
  \bibinfo{pages}{243--254}.
\newblock


\bibitem[\protect\citeauthoryear{Han, Mao, and Dally}{Han
  et~al\mbox{.}}{2015}]%
        {han2015deep}
\bibfield{author}{\bibinfo{person}{Song Han}, \bibinfo{person}{Huizi Mao},
  {and} \bibinfo{person}{William~J Dally}.} \bibinfo{year}{2015}\natexlab{}.
\newblock \showarticletitle{Deep compression: Compressing deep neural networks
  with pruning, trained quantization and huffman coding}.
\newblock \bibinfo{journal}{\emph{arXiv preprint arXiv:1510.00149}}
  (\bibinfo{year}{2015}).
\newblock


\bibitem[\protect\citeauthoryear{Hinton, Vinyals, and Dean}{Hinton
  et~al\mbox{.}}{2015}]%
        {hinton2015distilling}
\bibfield{author}{\bibinfo{person}{Geoffrey Hinton}, \bibinfo{person}{Oriol
  Vinyals}, {and} \bibinfo{person}{Jeff Dean}.}
  \bibinfo{year}{2015}\natexlab{}.
\newblock \showarticletitle{Distilling the knowledge in a neural network}.
\newblock \bibinfo{journal}{\emph{arXiv preprint arXiv:1503.02531}}
  (\bibinfo{year}{2015}).
\newblock


\bibitem[\protect\citeauthoryear{Howard, Zhu, Chen, Kalenichenko, Wang, Weyand,
  Andreetto, and Adam}{Howard et~al\mbox{.}}{2017}]%
        {howard2017mobilenets}
\bibfield{author}{\bibinfo{person}{Andrew~G Howard}, \bibinfo{person}{Menglong
  Zhu}, \bibinfo{person}{Bo Chen}, \bibinfo{person}{Dmitry Kalenichenko},
  \bibinfo{person}{Weijun Wang}, \bibinfo{person}{Tobias Weyand},
  \bibinfo{person}{Marco Andreetto}, {and} \bibinfo{person}{Hartwig Adam}.}
  \bibinfo{year}{2017}\natexlab{}.
\newblock \showarticletitle{Mobilenets: Efficient convolutional neural networks
  for mobile vision applications}.
\newblock \bibinfo{journal}{\emph{arXiv preprint arXiv:1704.04861}}
  (\bibinfo{year}{2017}).
\newblock


\bibitem[\protect\citeauthoryear{Huang, Chen, Li, Wu, van~der Maaten, and
  Weinberger}{Huang et~al\mbox{.}}{2017a}]%
        {huang2017multi}
\bibfield{author}{\bibinfo{person}{Gao Huang}, \bibinfo{person}{Danlu Chen},
  \bibinfo{person}{Tianhong Li}, \bibinfo{person}{Felix Wu},
  \bibinfo{person}{Laurens van~der Maaten}, {and} \bibinfo{person}{Kilian~Q
  Weinberger}.} \bibinfo{year}{2017}\natexlab{a}.
\newblock \showarticletitle{Multi-scale dense convolutional networks for
  efficient prediction}.
\newblock \bibinfo{journal}{\emph{arXiv preprint arXiv:1703.09844}}
  (\bibinfo{year}{2017}).
\newblock


\bibitem[\protect\citeauthoryear{Huang, Liu, van~der Maaten, and
  Weinberger}{Huang et~al\mbox{.}}{2017b}]%
        {huang2017condensenet}
\bibfield{author}{\bibinfo{person}{Gao Huang}, \bibinfo{person}{Shichen Liu},
  \bibinfo{person}{Laurens van~der Maaten}, {and} \bibinfo{person}{Kilian~Q
  Weinberger}.} \bibinfo{year}{2017}\natexlab{b}.
\newblock \showarticletitle{CondenseNet: An Efficient DenseNet using Learned
  Group Convolutions}.
\newblock \bibinfo{journal}{\emph{arXiv preprint arXiv:1711.09224}}
  (\bibinfo{year}{2017}).
\newblock


\bibitem[\protect\citeauthoryear{Huang, Liu, Weinberger, and van~der
  Maaten}{Huang et~al\mbox{.}}{2017c}]%
        {huang2017densely}
\bibfield{author}{\bibinfo{person}{Gao Huang}, \bibinfo{person}{Zhuang Liu},
  \bibinfo{person}{Kilian~Q Weinberger}, {and} \bibinfo{person}{Laurens van~der
  Maaten}.} \bibinfo{year}{2017}\natexlab{c}.
\newblock \showarticletitle{Densely connected convolutional networks}. In
  \bibinfo{booktitle}{\emph{Proceedings of the IEEE conference on computer
  vision and pattern recognition}}, Vol.~\bibinfo{volume}{1}.
  \bibinfo{pages}{3}.
\newblock


\bibitem[\protect\citeauthoryear{Hubara, Courbariaux, Soudry, El-Yaniv, and
  Bengio}{Hubara et~al\mbox{.}}{2016}]%
        {hubara2016quantized}
\bibfield{author}{\bibinfo{person}{Itay Hubara}, \bibinfo{person}{Matthieu
  Courbariaux}, \bibinfo{person}{Daniel Soudry}, \bibinfo{person}{Ran
  El-Yaniv}, {and} \bibinfo{person}{Yoshua Bengio}.}
  \bibinfo{year}{2016}\natexlab{}.
\newblock \showarticletitle{Quantized neural networks: Training neural networks
  with low precision weights and activations}.
\newblock \bibinfo{journal}{\emph{arXiv preprint arXiv:1609.07061}}
  (\bibinfo{year}{2016}).
\newblock


\bibitem[\protect\citeauthoryear{Iandola, Han, Moskewicz, Ashraf, Dally, and
  Keutzer}{Iandola et~al\mbox{.}}{2016}]%
        {iandola2016squeezenet}
\bibfield{author}{\bibinfo{person}{Forrest~N Iandola}, \bibinfo{person}{Song
  Han}, \bibinfo{person}{Matthew~W Moskewicz}, \bibinfo{person}{Khalid Ashraf},
  \bibinfo{person}{William~J Dally}, {and} \bibinfo{person}{Kurt Keutzer}.}
  \bibinfo{year}{2016}\natexlab{}.
\newblock \showarticletitle{SqueezeNet: AlexNet-level accuracy with 50x fewer
  parameters and< 0.5 MB model size}.
\newblock \bibinfo{journal}{\emph{arXiv preprint arXiv:1602.07360}}
  (\bibinfo{year}{2016}).
\newblock


\bibitem[\protect\citeauthoryear{Jia, Shelhamer, Donahue, Karayev, Long,
  Girshick, Guadarrama, and Darrell}{Jia et~al\mbox{.}}{2014}]%
        {jia2014caffe}
\bibfield{author}{\bibinfo{person}{Yangqing Jia}, \bibinfo{person}{Evan
  Shelhamer}, \bibinfo{person}{Jeff Donahue}, \bibinfo{person}{Sergey Karayev},
  \bibinfo{person}{Jonathan Long}, \bibinfo{person}{Ross Girshick},
  \bibinfo{person}{Sergio Guadarrama}, {and} \bibinfo{person}{Trevor Darrell}.}
  \bibinfo{year}{2014}\natexlab{}.
\newblock \showarticletitle{Caffe: Convolutional architecture for fast feature
  embedding}. In \bibinfo{booktitle}{\emph{Proceedings of the 22nd ACM
  international conference on Multimedia}}. ACM, \bibinfo{pages}{675--678}.
\newblock


\bibitem[\protect\citeauthoryear{Jouppi, Young, Patil, Patterson, Agrawal,
  Bajwa, Bates, Bhatia, Boden, Borchers, et~al\mbox{.}}{Jouppi
  et~al\mbox{.}}{2017}]%
        {jouppi2017datacenter}
\bibfield{author}{\bibinfo{person}{Norman~P Jouppi}, \bibinfo{person}{Cliff
  Young}, \bibinfo{person}{Nishant Patil}, \bibinfo{person}{David Patterson},
  \bibinfo{person}{Gaurav Agrawal}, \bibinfo{person}{Raminder Bajwa},
  \bibinfo{person}{Sarah Bates}, \bibinfo{person}{Suresh Bhatia},
  \bibinfo{person}{Nan Boden}, \bibinfo{person}{Al Borchers}, {et~al\mbox{.}}}
  \bibinfo{year}{2017}\natexlab{}.
\newblock \showarticletitle{In-datacenter performance analysis of a tensor
  processing unit}. In \bibinfo{booktitle}{\emph{Proceedings of the 44th Annual
  International Symposium on Computer Architecture}}. ACM,
  \bibinfo{pages}{1--12}.
\newblock


\bibitem[\protect\citeauthoryear{Krizhevsky, Nair, and Hinton}{Krizhevsky
  et~al\mbox{.}}{2014}]%
        {krizhevsky2014cifar}
\bibfield{author}{\bibinfo{person}{Alex Krizhevsky}, \bibinfo{person}{Vinod
  Nair}, {and} \bibinfo{person}{Geoffrey Hinton}.}
  \bibinfo{year}{2014}\natexlab{}.
\newblock \showarticletitle{The CIFAR-10 dataset}.
\newblock \bibinfo{journal}{\emph{online:
  https://www.cs.toronto.edu/~kriz/cifar.html}} (\bibinfo{year}{2014}).
\newblock


\bibitem[\protect\citeauthoryear{Krizhevsky, Sutskever, and Hinton}{Krizhevsky
  et~al\mbox{.}}{2012}]%
        {krizhevsky2012imagenet}
\bibfield{author}{\bibinfo{person}{Alex Krizhevsky}, \bibinfo{person}{Ilya
  Sutskever}, {and} \bibinfo{person}{Geoffrey~E Hinton}.}
  \bibinfo{year}{2012}\natexlab{}.
\newblock \showarticletitle{Imagenet classification with deep convolutional
  neural networks}. In \bibinfo{booktitle}{\emph{Advances in neural information
  processing systems}}. \bibinfo{pages}{1097--1105}.
\newblock


\bibitem[\protect\citeauthoryear{Latifi~Oskouei, Golestani, Hashemi, and
  Ghiasi}{Latifi~Oskouei et~al\mbox{.}}{2016}]%
        {latifi2016cnndroid}
\bibfield{author}{\bibinfo{person}{Seyyed~Salar Latifi~Oskouei},
  \bibinfo{person}{Hossein Golestani}, \bibinfo{person}{Matin Hashemi}, {and}
  \bibinfo{person}{Soheil Ghiasi}.} \bibinfo{year}{2016}\natexlab{}.
\newblock \showarticletitle{Cnndroid: Gpu-accelerated execution of trained deep
  convolutional neural networks on android}. In
  \bibinfo{booktitle}{\emph{Proceedings of the 2016 ACM on Multimedia
  Conference}}. ACM, \bibinfo{pages}{1201--1205}.
\newblock


\bibitem[\protect\citeauthoryear{LeCun}{LeCun}{1998}]%
        {lecun1998mnist}
\bibfield{author}{\bibinfo{person}{Yann LeCun}.}
  \bibinfo{year}{1998}\natexlab{}.
\newblock \showarticletitle{The MNIST database of handwritten digits}.
\newblock \bibinfo{journal}{\emph{http://yann. lecun. com/exdb/mnist/}}
  (\bibinfo{year}{1998}).
\newblock


\bibitem[\protect\citeauthoryear{Motamedi, Fong, and Ghiasi}{Motamedi
  et~al\mbox{.}}{2017}]%
        {motamedi2017machine}
\bibfield{author}{\bibinfo{person}{Mohammad Motamedi}, \bibinfo{person}{Daniel
  Fong}, {and} \bibinfo{person}{Soheil Ghiasi}.}
  \bibinfo{year}{2017}\natexlab{}.
\newblock \showarticletitle{Machine Intelligence on Resource-Constrained IoT
  Devices: The Case of Thread Granularity Optimization for CNN Inference}.
\newblock \bibinfo{journal}{\emph{ACM Transactions on Embedded Computing
  Systems (TECS)}} \bibinfo{volume}{16}, \bibinfo{number}{5s}
  (\bibinfo{year}{2017}), \bibinfo{pages}{151}.
\newblock


\bibitem[\protect\citeauthoryear{Motamedi, Fong, and Ghiasi}{Motamedi
  et~al\mbox{.}}{2018}]%
        {motamedi2017cappuccino}
\bibfield{author}{\bibinfo{person}{Mohammad Motamedi}, \bibinfo{person}{Daniel
  Fong}, {and} \bibinfo{person}{Soheil Ghiasi}.}
  \bibinfo{year}{2018}\natexlab{}.
\newblock \showarticletitle{Cappuccino: Efficient CNN Inference Software
  Synthesis for Mobile System-on-Chips}.
\newblock \bibinfo{journal}{\emph{IEEE Embedded Systems Letters}}
  (\bibinfo{year}{2018}).
\newblock


\bibitem[\protect\citeauthoryear{Ovtcharov, Ruwase, Kim, Fowers, Strauss, and
  Chung}{Ovtcharov et~al\mbox{.}}{2015}]%
        {ovtcharov2015accelerating}
\bibfield{author}{\bibinfo{person}{Kalin Ovtcharov}, \bibinfo{person}{Olatunji
  Ruwase}, \bibinfo{person}{Joo-Young Kim}, \bibinfo{person}{Jeremy Fowers},
  \bibinfo{person}{Karin Strauss}, {and} \bibinfo{person}{Eric~S Chung}.}
  \bibinfo{year}{2015}\natexlab{}.
\newblock \showarticletitle{Accelerating deep convolutional neural networks
  using specialized hardware}.
\newblock \bibinfo{journal}{\emph{Microsoft Research Whitepaper}}
  \bibinfo{volume}{2}, \bibinfo{number}{11} (\bibinfo{year}{2015}).
\newblock


\bibitem[\protect\citeauthoryear{Qiu, Wang, Yao, Guo, Li, Zhou, Yu, Tang, Xu,
  Song, et~al\mbox{.}}{Qiu et~al\mbox{.}}{2016}]%
        {qiu2016going}
\bibfield{author}{\bibinfo{person}{Jiantao Qiu}, \bibinfo{person}{Jie Wang},
  \bibinfo{person}{Song Yao}, \bibinfo{person}{Kaiyuan Guo},
  \bibinfo{person}{Boxun Li}, \bibinfo{person}{Erjin Zhou},
  \bibinfo{person}{Jincheng Yu}, \bibinfo{person}{Tianqi Tang},
  \bibinfo{person}{Ningyi Xu}, \bibinfo{person}{Sen Song}, {et~al\mbox{.}}}
  \bibinfo{year}{2016}\natexlab{}.
\newblock \showarticletitle{Going deeper with embedded fpga platform for
  convolutional neural network}. In \bibinfo{booktitle}{\emph{Proceedings of
  the 2016 ACM/SIGDA International Symposium on Field-Programmable Gate
  Arrays}}. ACM, \bibinfo{pages}{26--35}.
\newblock


\bibitem[\protect\citeauthoryear{Russakovsky, Deng, Su, Krause, Satheesh, Ma,
  Huang, Karpathy, Khosla, Bernstein, Berg, and Fei-Fei}{Russakovsky
  et~al\mbox{.}}{2015}]%
        {ILSVRC15}
\bibfield{author}{\bibinfo{person}{Olga Russakovsky}, \bibinfo{person}{Jia
  Deng}, \bibinfo{person}{Hao Su}, \bibinfo{person}{Jonathan Krause},
  \bibinfo{person}{Sanjeev Satheesh}, \bibinfo{person}{Sean Ma},
  \bibinfo{person}{Zhiheng Huang}, \bibinfo{person}{Andrej Karpathy},
  \bibinfo{person}{Aditya Khosla}, \bibinfo{person}{Michael Bernstein},
  \bibinfo{person}{Alexander~C. Berg}, {and} \bibinfo{person}{Li Fei-Fei}.}
  \bibinfo{year}{2015}\natexlab{}.
\newblock \showarticletitle{{ImageNet Large Scale Visual Recognition
  Challenge}}.
\newblock \bibinfo{journal}{\emph{International Journal of Computer Vision
  (IJCV)}} \bibinfo{volume}{115}, \bibinfo{number}{3} (\bibinfo{year}{2015}),
  \bibinfo{pages}{211--252}.
\newblock
\urldef\tempurl%
\url{https://doi.org/10.1007/s11263-015-0816-y}
\showDOI{\tempurl}


\bibitem[\protect\citeauthoryear{Shazeer, Mirhoseini, Maziarz, Davis, Le,
  Hinton, and Dean}{Shazeer et~al\mbox{.}}{2017}]%
        {shazeer2017outrageously}
\bibfield{author}{\bibinfo{person}{Noam Shazeer}, \bibinfo{person}{Azalia
  Mirhoseini}, \bibinfo{person}{Krzysztof Maziarz}, \bibinfo{person}{Andy
  Davis}, \bibinfo{person}{Quoc Le}, \bibinfo{person}{Geoffrey Hinton}, {and}
  \bibinfo{person}{Jeff Dean}.} \bibinfo{year}{2017}\natexlab{}.
\newblock \showarticletitle{Outrageously large neural networks: The
  sparsely-gated mixture-of-experts layer}.
\newblock \bibinfo{journal}{\emph{arXiv preprint arXiv:1701.06538}}
  (\bibinfo{year}{2017}).
\newblock


\bibitem[\protect\citeauthoryear{Szegedy, Liu, Jia, Sermanet, Reed, Anguelov,
  Erhan, Vanhoucke, Rabinovich, et~al\mbox{.}}{Szegedy et~al\mbox{.}}{2015}]%
        {szegedy2015going}
\bibfield{author}{\bibinfo{person}{Christian Szegedy}, \bibinfo{person}{Wei
  Liu}, \bibinfo{person}{Yangqing Jia}, \bibinfo{person}{Pierre Sermanet},
  \bibinfo{person}{Scott Reed}, \bibinfo{person}{Dragomir Anguelov},
  \bibinfo{person}{Dumitru Erhan}, \bibinfo{person}{Vincent Vanhoucke},
  \bibinfo{person}{Andrew Rabinovich}, {et~al\mbox{.}}}
  \bibinfo{year}{2015}\natexlab{}.
\newblock \showarticletitle{Going deeper with convolutions}. Cvpr.
\newblock


\bibitem[\protect\citeauthoryear{Wu, Wan, Yue, and Keutzer}{Wu
  et~al\mbox{.}}{2017}]%
        {wu2017squeezeseg}
\bibfield{author}{\bibinfo{person}{Bichen Wu}, \bibinfo{person}{Alvin Wan},
  \bibinfo{person}{Xiangyu Yue}, {and} \bibinfo{person}{Kurt Keutzer}.}
  \bibinfo{year}{2017}\natexlab{}.
\newblock \showarticletitle{Squeezeseg: Convolutional neural nets with
  recurrent crf for real-time road-object segmentation from 3d lidar point
  cloud}.
\newblock \bibinfo{journal}{\emph{arXiv preprint arXiv:1710.07368}}
  (\bibinfo{year}{2017}).
\newblock


\bibitem[\protect\citeauthoryear{Zeiler and Fergus}{Zeiler and Fergus}{2014}]%
        {zeiler2014visualizing}
\bibfield{author}{\bibinfo{person}{Matthew~D Zeiler} {and} \bibinfo{person}{Rob
  Fergus}.} \bibinfo{year}{2014}\natexlab{}.
\newblock \showarticletitle{Visualizing and understanding convolutional
  networks}. In \bibinfo{booktitle}{\emph{European conference on computer
  vision}}. Springer, \bibinfo{pages}{818--833}.
\newblock


\bibitem[\protect\citeauthoryear{Zhang, Li, Sun, Guan, Xiao, and Cong}{Zhang
  et~al\mbox{.}}{2015}]%
        {zhang2015optimizing}
\bibfield{author}{\bibinfo{person}{Chen Zhang}, \bibinfo{person}{Peng Li},
  \bibinfo{person}{Guangyu Sun}, \bibinfo{person}{Yijin Guan},
  \bibinfo{person}{Bingjun Xiao}, {and} \bibinfo{person}{Jason Cong}.}
  \bibinfo{year}{2015}\natexlab{}.
\newblock \showarticletitle{Optimizing fpga-based accelerator design for deep
  convolutional neural networks}. In \bibinfo{booktitle}{\emph{Proceedings of
  the 2015 ACM/SIGDA International Symposium on Field-Programmable Gate
  Arrays}}. ACM, \bibinfo{pages}{161--170}.
\newblock


\end{thebibliography}
\end{document}